\documentclass{article}

\usepackage{times}
\usepackage{graphicx} 
\usepackage{subfigure} 

\usepackage{natbib}

\usepackage{algorithm}
\usepackage{algorithmic}

\usepackage{amssymb}
\usepackage{amsmath}
\usepackage{amsfonts}
\usepackage{latexsym}
\usepackage{booktabs}
\usepackage{siunitx}

\usepackage{hyperref}

\usepackage[accepted]{icml2017}

\icmltitlerunning{Improved Representation Learning for Predicting Commonsense Ontologies}

\begin{document} 

\twocolumn[
\icmltitle{Improved Representation Learning for Predicting Commonsense Ontologies}

\icmlsetsymbol{equal}{*}

\begin{icmlauthorlist}
\icmlauthor{Xiang Li}{umass}
\icmlauthor{Luke Vilnis}{umass}
\icmlauthor{Andrew McCallum}{umass}
\end{icmlauthorlist}

\icmlaffiliation{umass}{CICS, University of Massachusetts, Amherst, USA}

\icmlcorrespondingauthor{Xiang Li}{xiangl@cs.umass.edu}

\vskip 0.3in
]

\printAffiliationsAndNotice{}

\begin{abstract}
Recent work in learning ontologies (hierarchical and partially-ordered structures) has leveraged the intrinsic geometry of spaces of learned representations to make predictions that automatically obey complex structural constraints. We explore two extensions of one such model, the order-embedding \cite{orderembedding} model for hierarchical relation learning, with an aim towards improved performance on text data for commonsense knowledge representation. Our first model jointly learns ordering relations and non-hierarchical knowledge in the form of raw text. Our second extension exploits the partial order structure of the training data to find long-distance triplet constraints among embeddings which are poorly enforced by the pairwise training procedure. We find that both incorporating free text and augmented training constraints improve over the original order-embedding model and other strong baselines.
\end{abstract} 

\section{Introduction}

A core problem in artificial intelligence is to capture, in machine-usable form, the collection of information that an ordinary person would have, known as \emph{commonsense knowledge}. For example, a machine should know that a room may have a door, and that when a person enters a room, it is generally through a door. This background knowledge is crucial for solving many difficult, ambiguous natural language problems in coreference resolution and question answering, as well as the creation of other reasoning machines.

More than just curating a static collection of facts, we would like commonsense knowledge to be represented in a way that lends itself to machine reasoning and inference of missing information. We concern ourselves in this paper with the problem of learning commonsense knowledge representations.

In machine learning settings, knowledge is usually represented as a hypergraph of triplets such as Freebase~\cite{bollacker2008freebase}, WordNet ~\cite{miller1995wordnet}, and ConceptNet ~\cite{speer2016conceptnet}. In these knowledge graphs, nodes represent entities or terms $t$, and hyperedges are relations $R$ between these entities or terms, with each fact in the knowledge graph represented as a triplet $<t_1, R, t_2>$. Researchers have developed many models for knowledge representation and learning in this setting~\cite{transe, transh, bilinear2011, li2016commonsense, socher2013reasoning}, under the umbrella of \emph{knowledge graph completion}. However, none of these naturally lend themselves to traditional methods of logical reasoning such as transitivity and negation.

While a knowledge graph completion model can represent relations such as {\sc Is-A} and entailment, there is no mechanism to ensure that its predictions are internally consistent. For example, if we know that a dog is a mammal, and a pit bull is a dog, we would like the model to also predict that a pit bull is a mammal. These transitive entailment relations describe ontologies of hierarchical data, a key component of commonsense knowledge which we focus on in this work.

Recently, a thread of research on representation learning has aimed to create embedding spaces that automatically enforce consistency in these predictions using the intrinsic geometry of the embedding space \cite{vilnis2014word,orderembedding,DBLP:journals/corr/NickelK17}. In these models, the inferred embedding space creates a globally consistent structured prediction of the ontology, rather than the local relation predictions of previous models.

We focus on the order-embedding model \cite{orderembedding} which was proposed for general hierarchical prediction including multimodal problems such as image captioning. While the original work included results on ontology prediction on WordNet, we focus exclusively on the model's application to commonsense knowledge, with its unique characteristics including complex ordering structure, compositional, multi-word entities, and the wealth of commonsense knowledge to be found in large-scale unstructured text data.

We propose two extensions to the order embedding model. The first augments hierarchical supervision from existing ontologies with non-hierarchical knowledge in the form of raw text. We find incorporating unstructured text brings accuracy from 92.0 to 93.0 on a commonsense dataset containing {\sc Is-A} relations from ConceptNet and Microsoft Concept Graph (MCG), with larger relative gains from smaller amounts of labeled data.

The second extension uses the complex partial-order structure of real-world ontologies to find long-distance triplet constraints among embeddings which are poorly enforced by the standard pairwise training method.  By adding our additional triplet constraints to the baseline order-embedding model, we find performance improves from 90.6 to 91.3 accuracy on the WordNet ontology dataset.

We find that order embeddings' ease of extension, both by incorporating non-ordered data, and additional training constraints derived from the structure of the problem, makes it a promising avenue for the development of further algorithms for automatic learning and jointly consistent prediction of ontologies.

\section{Data}

In this work, we use the ConceptNet \cite{speer2016conceptnet}, WordNet \cite{miller1995wordnet}, and Microsoft Concept Graph (MCG) \cite{wu2012probase, wang2015inference} knowledge bases for our ontology prediction experiments.

WordNet is a knowledge base (KB) of single words and relations between them such as hypernymy and meronymy. For our task, we use the hypernym relations only. ConceptNet is a KB of triples consisting of a left term $t_1$, a relation $R$, and a right term $t_2$. The relations come from a fixed set of size 34. But unlike WordNet, terms in ConceptNet can be phrases. We focus on the {\sc Is-A} relation in this work. MCG also consists of hierarchical relations between multi-word phrases, ranging from extremely general to specific. Examples from each dataset are shown in Table \ref{table:data_examples}. 

For experiments involving unstructured text, we use the WaCkypedia corpus \cite{baroni2009wacky}.

\begin{table*}[t]
\centering
\small
\begin{tabular}{|ccc|c|}
\hline
term 1 & relation & term 2 & Dataset\\\hline
coral reefs & IsA & delicate ecosystems & ConceptNet\\
diabetes & IsA & chronic health condition & MCG\\
fantasy\_life.n.01 & IsA & imagination.n.01 & WordNet\\
\hline
\end{tabular}
\caption{Example triplets from each dataset.}
\label{table:data_examples}
\end{table*}

\section{Models}

We introduce two variants of order embeddings. The first incorporates non-hierarchical unstructured text data into the supervised ontology. The second improves the training procedure by adding additional examples representing long-range constraints.

\subsection{Order Embeddings}

Order Embeddings are a model for automatically enforcing partial-ordering (or lattice) constraints among predictions directly in embedding space. The vector embeddings satisfy the following property with respect to the partial order:
\begin{align*}
x \preceq y  \text{ if and only if } \bigwedge _{i=1}^{N}x_{i}\geq y_i
\end{align*}
\begin{figure}[t]
\begin{center}
\includegraphics[width=6cm, height=5cm ]{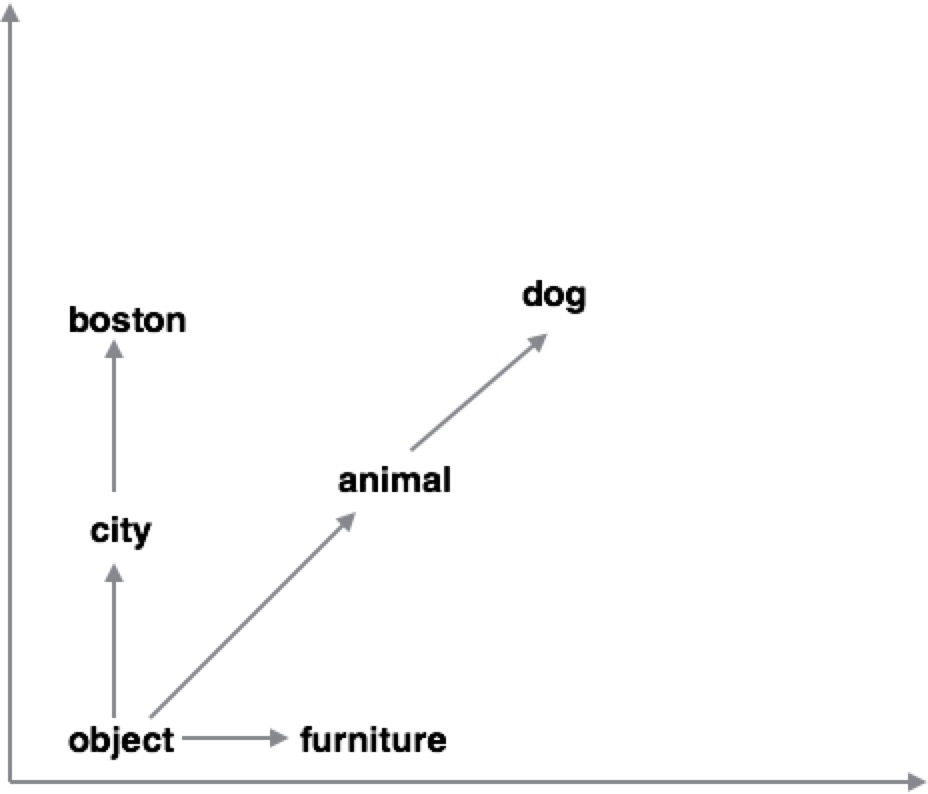}
\caption{Order Embedding}\label{oe}
\end{center}
\end{figure}

where $x$ is the subcategory and $y$ is the supercategory. This means the general concept embedding should be smaller than the specific concept embedding in every coordinate of the embeddings. An illustration of this geometry can be found in Figure 1. We can define a surrogate energy for this ordering function as $d(x, y) = \left \| \max(0,y-x) \right \|^2$. The learning objective for order embeddings becomes the following, where $m$ is a margin parameter, $x$ and $y$ are the hierarchically supervised pairs, and $x'$ and $y'$ are negatively sampled concepts:
\begin{align*}
L_{\text{Order}} = \sum_{x,y}\max(0, m+d(x,y)-d(x', y'))
\end{align*}

\subsection{Joint Text and Order Embedding}

We aim to augment our ontology prediction embedding model with more general commonsense knowledge mined from raw text. A standard method for learning word representations is word2vec \cite{mikolov2013distributed}, which predicts current word embeddings using a context of surrounding word embeddings. We incorporate a modification of the CBOW model in this work, which uses the average embedding from a window around the current word as a context vector $v_2$ to predict the current word vector $v_1$:
\begin{align*}
v_2 = \frac{1}{window}\sum_{k \in \{-window/2,...,window/2\}\setminus \{t\}}v_{t+k}
\end{align*}
Because order embeddings are all positive and compared coordinate-wise, we use a variant of CBOW that scores similarity to context based on based on $L_1$ distance and not dot product, $v'_1$ and $v'_2$ are the negative examples selected from the vocabulary during training:
\begin{align*}
& d_\text{pos} = d(v_1,v_2) = \left \| v_1- v_2\right \|\\
& d_\text{neg} = d(v'_1, v'_2) =  \left \| v'_1- v'_2\right \| \\
& L_{\text{CBOW}}= \sum_{w_c,w_t}\max(0, m+d_\text{pos}-d_\text{neg})
\end{align*}

Finally, after each gradient update, we map the embeddings back to the positive domain by applying the absolute value function. We propose jointly learning both the order- and text- embedding model with a simple weighted combination of the two objective functions:
\begin{align*}
&L_{\text{Joint}} = \alpha_{1}L_{\text{Order}}+\alpha_{2}L_{\text{CBOW}}
\end{align*}

\section{Long-Range Join and Meet Constraints}

\begin{figure}[t]
\begin{center}
\includegraphics[width=6cm, height=5cm ]{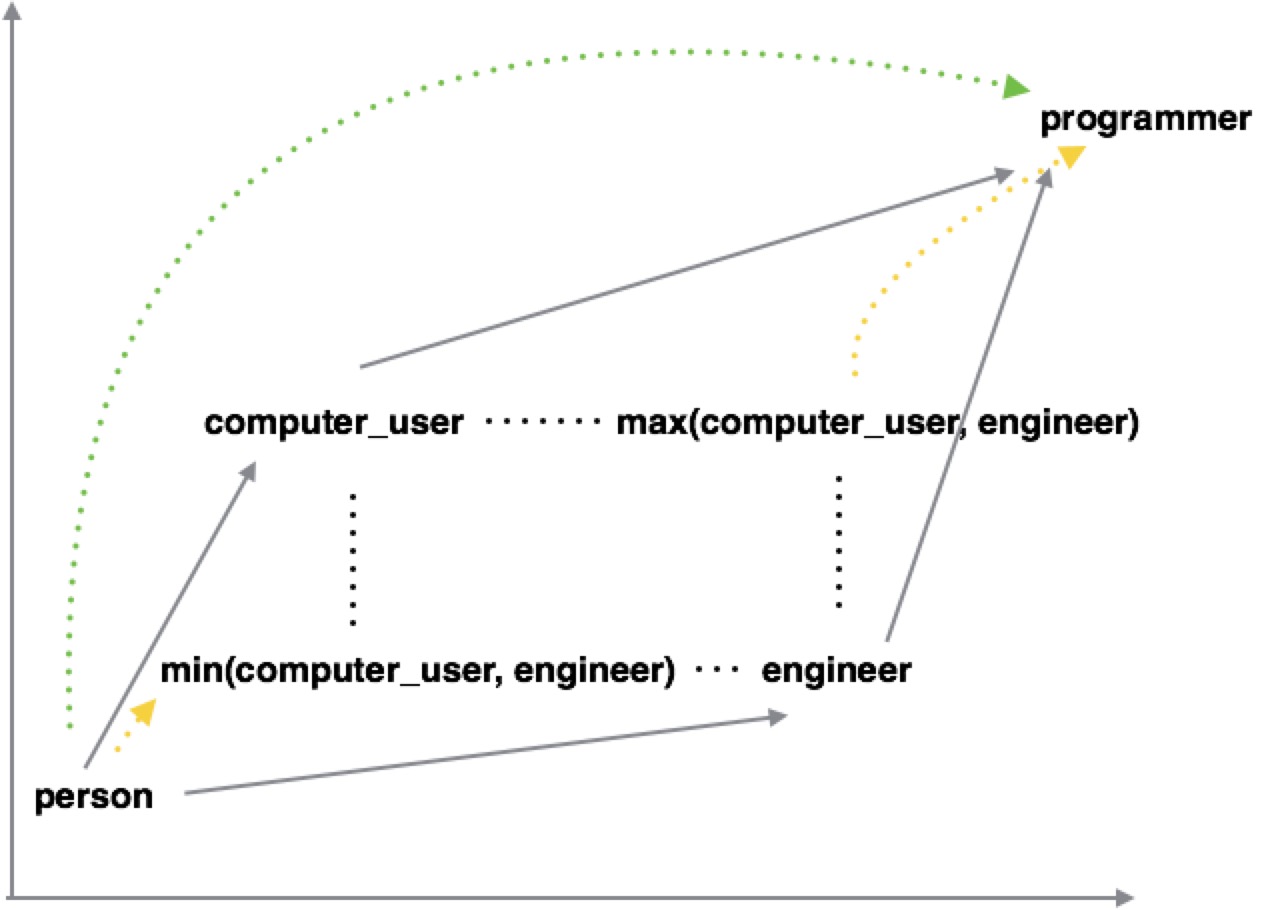}
\caption{Adding more training examples: black line is the original training data, green line is obtained by transitive closure, and yellow line is obtained by join and meet.}\label{g1}
\end{center}
\end{figure}

Order embeddings map words to a partially-ordered space, which we can think of as a directed acyclic graph (DAG). A simple way to add more training examples is to take the transitive closure of this graph. For example, if we have $<$dog IsA mammal$>$, $<$mammal IsA animal$>$, we can produce the training example $<$dog IsA animal$>$.

We observe that even more training examples can be created by treating our partial-order structure as a lattice. A lattice is a partial order equipped with two additional operations, \emph{join} and \emph{meet}. The join and meet of a pair P are respectively the supremum (least upper bound) of P, denoted $\vee$, and the infimum (greatest lower bound), denoted $\wedge$. In our case, the vector join and meet would be the pointwise max and min of two embeddings.

We can add many additional training examples to our data by enforcing that the vector join and meet operations satisfy the joins and meets found in the training lattice/DAG. If $w_c$ and $w_p$ are the nearest common child and parent for a pair $w_1, w_2$, the loss for join and meet learning can be written as the following:
\begin{align*}
& d_c(w_1,w_2,w_c) = \left \| \max(0,w_1 \vee w_2-w_c) \right \|^2 \\
& d_p(w_1,w_2,w_p) = \left \| \max(0,w_p - w_1 \wedge w_2) \right \|^2 \\
& {\small L_\text{join} = \sum_{w_1,w_2,w_c}\max(0, m+d_c(w_1,w_2,w_c))}\\
& {\small L_\text{meet} = \sum_{w_1,w_2,w_p}\max(0, m+d_p(w_1,w_2,w_p))}\\
& L = L_\text{join} + L_\text{meet}
\end{align*}

\section{Experiments}

In both sets of experiments we train all models using the Adam optimizer \cite{kingma2014adam}, using embeddings of dimension 50, with all hyperparameters tuned on a development set. When embedding multi-word phrases, we represent them as the average of the constituent word embeddings.

\subsection{Joint Text and Order Embedding}
We perform two sets of experiments on the combined ConceptNet and MCG {\sc Is-A} relations, using different amounts of training and testing data. The first data set, called Data1, uses 119,159 training examples, 1,089 dev examples, and 1,089 test examples. The second dataset, Data2, evenly splits the data in 47,662 examples for each set.

Our baselines for this model are a standard order embedding model, and a bilinear classifier \cite{bilinear2011} trained to predict {\sc Is-A}, both with and without additional unstructured text augmenting the model in the same way as the joint order embedding model.

We see in Table 2 that while adding extra text data helps all models, the best performance is consistently achieved by a combination of order embeddings and unstructured text.

\begin{table}[t]
\centering
\label{joint-text}
\begin{tabular}{l|l|l}

Model         & Data1 Acc& Data2 Acc \\ \hline
Bilinear      & 90.5           & 77.3          \\ 
OE            & 92.0           & 78.1          \\ 
Bilinear+Cbow & 92.4           & 80.1          \\ 
OE+Cbow       & \textbf{93.0}           & \textbf{80.4}         \\ 
\end{tabular}
\caption{Joint Text and Order Embedding}
\end{table}

\begin{table}[t]
\centering
\label{join-meet}
\begin{tabular}{l|l}

Model              & Accuracy \\ \hline
transitive closure & 88.2     \\
word2gauss         & 86.6     \\ 
OE                 & 90.6     \\ 
OE+Join \& Meet   & \textbf{91.3}   \\ 
\end{tabular}
\caption{Join and Meet Constraints}
\end{table}

\subsection{Long-Range Join and Meet Constraints}

In this experiment, we use the same dataset as \cite{orderembedding}, created by taking 40,00 edges from the 838,073-edge transitive closure of the WordNet hierarchy for the dev set, 4,000 for the test set, and training on the rest of the transitive closure. We additionally add the long-range join and meet constraints (3,028,302 and 4,006 respectively) between different concepts and see that the inclusion of this additional supervision results in further improvement over the baseline order embedding model, as seen in Table 3.

\section{Conclusion and Future Work}

In this work we presented two extensions to the order embedding model. The first incorporates unstructured text to improve performance on {\sc Is-A} relations, while the second uses long-range constraints automatically derived from the ontology to provide the model with more useful global supervision. In future work we would like to explore embedding models for structured prediction that automatically incorporate additional forms of reasoning such as negation, joint learning of ontological and other commonsense relations, and the application of improved training methods to new models for ontology prediction such as Poincar{\' e} embeddings.

\bibliography{example_paper}
\bibliographystyle{icml2017}

\end{document}